\documentclass[conference]{IEEEtran}
\usepackage{cite}
\usepackage{graphicx}
\graphicspath{ {images/} }
\IEEEoverridecommandlockouts
\usepackage{cite}
\usepackage{amsmath,amssymb,amsfonts,bbm}
\usepackage{subfigure,bm,balance}
\usepackage[ruled,vlined]{algorithm2e}
\usepackage{graphicx}
\usepackage{textcomp}
\usepackage{xcolor}
\begin{document}


\title{FiSTECH: Financial Style Transfer to Enhance Creativity without Hallucinations in LLMs}
\author{\IEEEauthorblockN{Sohini Roychowdhury,
Marko Krema,
Brian Moore,
Xingjian Lai,
Dike Effedua,
Bharat Jethwani
}
\IEEEauthorblockA{Corporate Data and Analytics Office (CDAO), Accenture LLP, USA, Email: sohini.roychowdhury@accenture.com}}

\maketitle
\begin{abstract}
Recent trends in Generative AI have emerged towards fine-tuning foundational large language models (LLMs) to create domain-specific LLMs for automation and chatbot-like applications. Specialized applications for analytics-heavy domains such as \textit{Financial report generation} require specific writing styles that comprise compound and creative sentences with minimized hallucinations. In this work, we explore the self-corrective auto-regressive qualities of LLMs to learn creativity in writing styles with minimal prompting. We propose a novel two-stage fine-tuning (FT) strategy wherein in the first stage public domain financial reports are used to train for writing styles while allowing the LLM to hallucinate. In the second stage the examples of hallucinations are manually corrected and further used to fine-tune the LLM. The finally trained LLM learns to generate specific financial report sections using minimal instructions and tabular data inputs while ensuring low fine-tuning costs. Our proposed two-stage fine-tuning boosts the accuracy of financial questions answering by two-folds while reducing hallucinations by over 50\%. Also, the fine-tuned model has lower perplexity, improved ROUGE, TER and BLEU scores, higher creativity and knowledge density with lower uncertainty and cross entropy than base LLMs. Thus, the proposed framework can be generalized to train creativity in LLMs by first allowing them to hallucinate.
\end{abstract}
\section{Introduction}\label{intro}
Large language models (LLMs) have powered several question answering chat-bots and automation applications as major use-cases in the recent past. While most research advancements have been around foundational LLMs \cite{genllms} such as ChatGPTs, LLama, Gemini, Claude etc., domain specific products have typically benefited largely from retrieval augmented generation approaches (RAG) \cite{bigdata} and limited training \cite{domainllms} \cite{fin2}. The \textit{financial domain} specifically, is characterized with significant numerical data, data transformations, abbreviations and definitions. Some recent industrial products and use cases in this domain include: automated financial statement analysis, personalized narrative generation for financial reports, automated tagging and labeling of financial data and reports, financial forecasting and prediction, risk management and compliance and audit processes \cite{fin1}. The major notable contributions for LLMs in the domain of finance include the BloombergGPT \cite{bloomberg} that is capable of sentiment analysis, named entity recognition, and question answering when applied to financial text; and Financial GPT\cite{fin2} that incorporates various financial data formats, including news, filings, social media, and company announcements into the training phase to enable creation of financial products and services and supports informed investment and risk management strategies.

\begin{figure*}[ht]
    \centering
    \subfigure[Text with highlighted named entities or KG nodes.]
	{\includegraphics[width=0.98\textwidth]{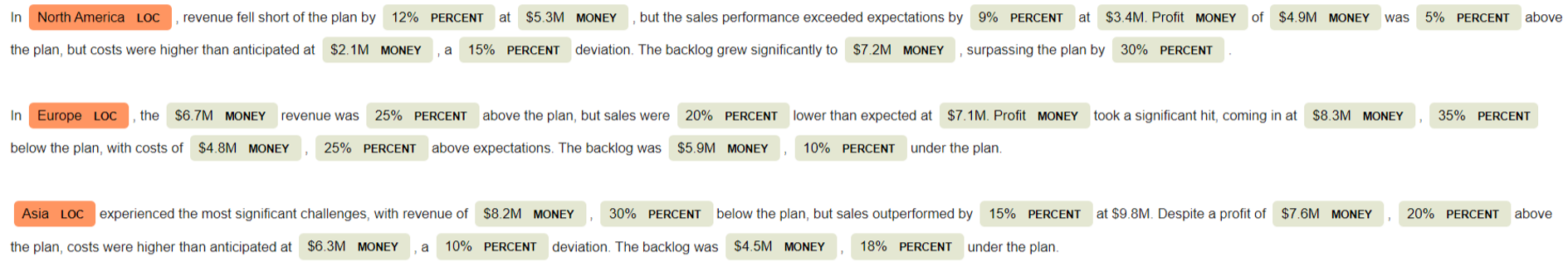}}
	\subfigure[KG source, destination and relationships extracted.]
	{\includegraphics[width=0.4\textwidth]{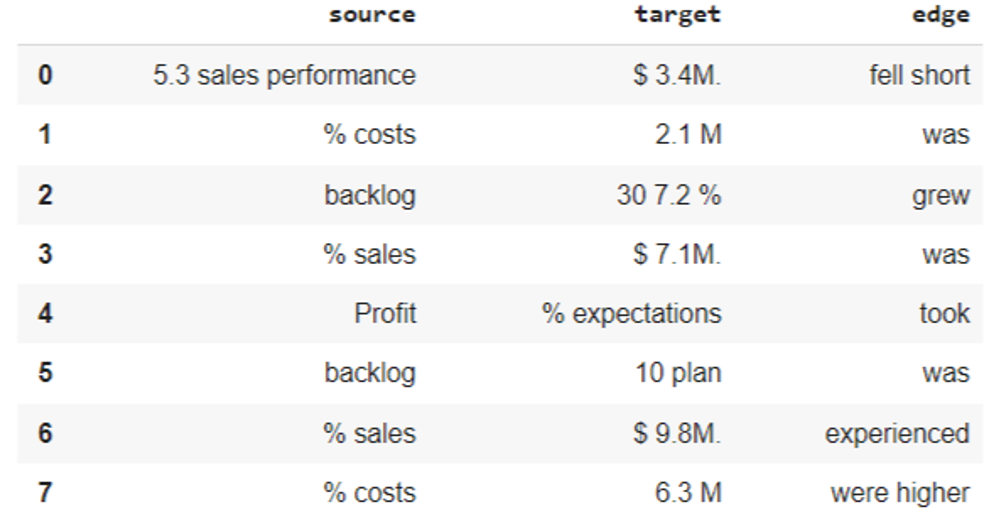}}
    \caption{An example of KG generation from financial text. First, the named entities are detected in (a) followed by extraction of the complete KG that captures the relationships between the nodes in (b). The most pertinent KG to a user-query is retrieved for financial applications.}
    \label{fig:kg}
\end{figure*}

While building financial-domain specific LLM applications poses challenges with regards to training data quality and hardware resources needed for large scale training epochs, most of these use-cases rely on web-searches combined with knowledge graphs (KGs) to ensure \textit{recency} in the data  and quality standards of responses/outputs \cite{fin1}. The generative AI-based approach of retrieving updated financial information and serving the analysis in domain-representative jargon, also known as the agentic-RAG, comprises two steps. First, a web-search is orchestrated for the user-query; second from the retrieved web-link texts, paragraphs that contain relevant information are identified using the KGs. Typically, KGs are a structured representation of data or textual information and they consist of nodes (named entities) and edges (relationships) that connect them. For example, a node might represent a named entity such as an organization, place, or person, while an edge could represent a relationship like \textit{resulted in}, \textit{risen/fallen} etc. An example of KG representation from financial data is shown in Fig. \ref{fig:kg}. The most appropriate paragraphs/KGs from the web links are then used to serve the information in a personalized and structured manner to create accurate and reliable AI systems that are capable of fact-checking, improved understanding, enhanced domain-specific reasoning \cite{comb1}. 

\begin{figure}[ht]
    \centering
    \includegraphics[width=3.2in]{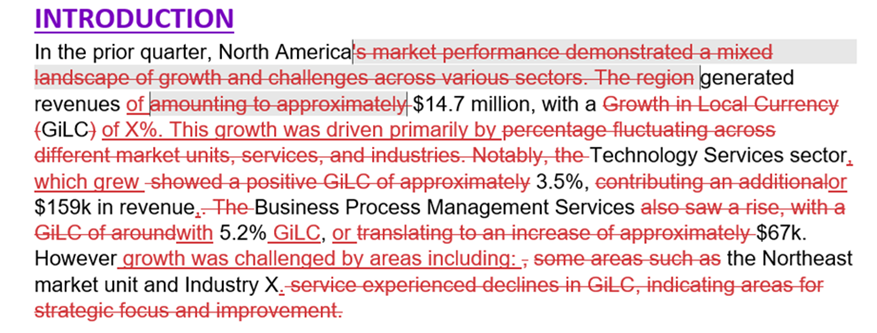}
    \caption{Example of simple RAG-based approach using GPT4o to generate Financial Reporting text using input data as instructions and tabular sources.}
    \label{fig:sys}
\end{figure}

Although agentic-RAG processes have benefited some financial use-cases \cite{genaiagent1}, these approaches do not scale to financial report generation tasks, wherein, the intention is to transfer the domain-specific writing style. As an experiment, we used tabular data and English language instructions describing the persona of a financial analyst and specific instructions to the GPT4o LLM to generate multiple paragraphs of financial reporting text. The sample output is shown in Fig. \ref{fig:sys}, where unwanted text is struck out by our financial domain experts. Advanced prompt engineering and instructional guardrails \cite{bigdata} for this use-case led to the following observations.
\begin{itemize}
\item General wordiness of the output and the use of unwanted words such as \{successively, landscape of growth\} did not change even after providing detailed instructions.
\item Controlling for LLM parameters like temperature, top\_p and max tokens did not improve the overall \textit{quality} of generated text.
\item Increased instructions to generate multiple sentences from tabular data reduced the overall performance of text generation. 
\item Compound sentences that are atypical for the financial domain, such as contrasting sentiments regarding the same entity being presented in a single compound sentence, cannot be generated reliably using the agentic-RAG approaches.
\end{itemize}

One major drawback of LLM-based solutions is the occurrence of \textit{fake responses} or \textit{hallucinations} as prompt responses \cite{bigdata}. \textit{Hallucinations} are largely caused by the uncertainty in the later layers for predicting the \textit{next token/word} in a response sequence \cite{hall1}. Existing works in \cite{hall1} \cite{hal2} have shown that hallucinations are unrepeatable occurrences caused by uncertainty in token generation and the same model parameters are noticed for \textit{creative} responses as well. The major difference between \textit{creativity} and \textit{hallucinations} is that creative responses are true facts that may not be present in the context provided to the LLM, but they may be learned or extrapolated accurately by the LLM. Hallucinations on the other hand are also contextual bifurcations from the knowledge available to the LLM, but they are factually inaccurate. In our prior works \cite{bigdata}, we have demonstrated hallucination control checkers that can be implemented per query-context-response level. In this work, we make a hypothesis: ``Intelligence requires the capability of learning from mistakes”. Thus, instead of fine-tuning LLMs while controlling for hallucination metrics, we demonstrate the self-correction capabilities of LLMs in two-stages based on the prior work in \cite{hall1}. This hypothesis is analogous to the process in which early learners/children learn a new skill/creativity by exploration and making mistakes followed by learning from their mistakes. 

\begin{figure}[ht]
    \centering
    \includegraphics[width=3.2in]{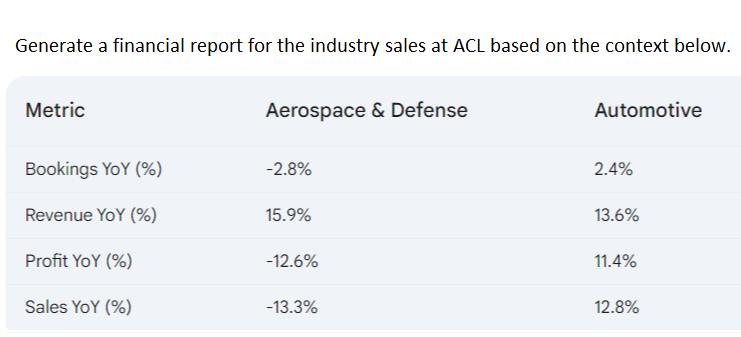}
    \caption{Basic prompt with instructions and tabular data as input.}
    \label{p1}
\end{figure}

\begin{figure*}[ht]
     \centering
    \subfigure[Examples of compound sentences in the generated output.]
	{\includegraphics[width=0.98\textwidth]{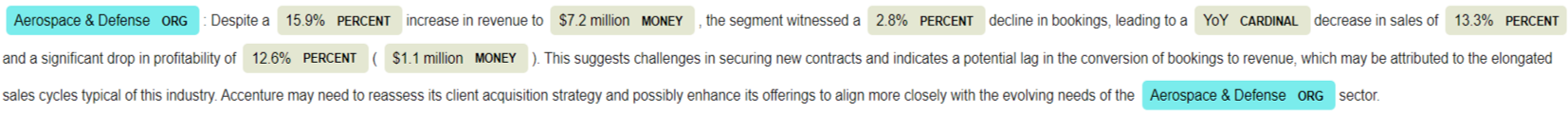}}
	\subfigure[Examples of creative sentences in the generated output.]
	{\includegraphics[width=0.98\textwidth]{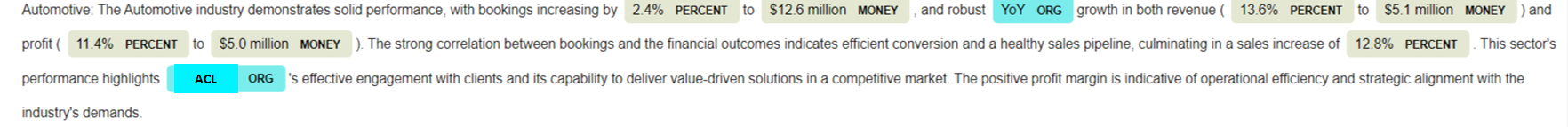}}
    \caption{Examples of two-stage fine tuned financial report generation paragraphs. The two major considerations in style-transfer are the formation of compound sentences in (a) and creative logical sentences that are not hallucinations in (b).}
    \label{fig:gr}
\end{figure*}

This paper makes two key contributions. First, we propose a two-stage LLM fine-tuning framework that enables creative and compound writing style transfer while minimizing domain-specific hallucinations. We demonstrate the step-wise enhancements in the knowledge density per generated paragraph across the fine-tuning stages, while ensuring minimal fine-tuning cost of under \$18 for fine-tuning the GPT3.5 model. Second, we propose multiple novel metrics that can assess the performance of fine-tuned LLMs that are based on KG-based approaches. These metrics enable tracking the required creativity standards per generated sentence while flagging hallucinations using ``spacy''-based libraries. We test the fine-tuned model by applying a basic prompt that includes minimal instructions and tabular data shown in Fig. \ref{p1} as input and the corresponding output with creative and compound sentences is shown in Fig. \ref{fig:gr}.

\section{Related Work}
\begin{figure}[ht]
    \centering
    \includegraphics[width=3.2in]{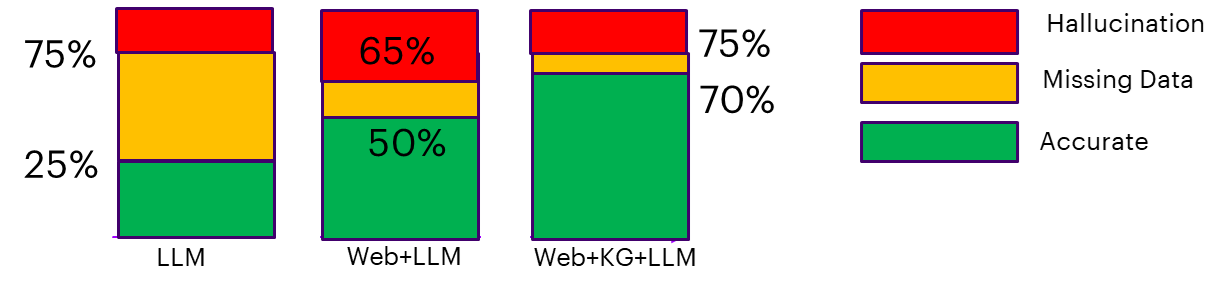}
    \caption{Impact of LLM augmentative approaches using LLM only, Web search$+$LLM and Web search$+$KG$+$LLM to improve quality of responses based on \cite{luna}.}\label{fig:q}
\end{figure}

Domain-specific generative AI led automation tasks such as a financial chatbot or financial news writer, have seen significant improvements in overall knowledge retrieval tasks by combining search engine and KG capabilities as shown in Fig. \ref{fig:q} \cite{luna}. The major reason for the improved response quality by adding on web searches and KG isolation of text is that the LLMs follow a unique knowledge distribution, with a head, body/torso and tail \cite{luna2}. Knowledge from the LLM head (commonly occurring and rarely changing facts) are easily retrievable with minimal hallucinations. Contrastingly, knowledge from the LLM tail (rapidly changing facts, such as share prices etc.) can lead to stale data in the responses or inaccuracies/hallucinations or ``I dont know'' responses from LLMs. Thus, it is imperative to teach the LLM to work with latest data and reasonably modify the text generation style to ensure lowered hallucinations and unknown responses as shown in \cite{luna2}. A summary of latest works on the financial domain, style-transfer and detection of hallucination and creativity are presented below.
\subsection{LLMs for the Financial Domain}
So far, LLMs have provided advanced capabilities for insights, trends, and assessments in the financial domain. Notable models like FINBERT \cite{table1-1}, introduced in 2022, demonstrate the adaptation of LLMs to financial domains. Innovations continue with Bloomberg GPT \cite{bloomberg} with 50 billion parameters trained on extensive financial domain data, making it one of the largest and most powerful financial-specific LLMs to date. Looking forward, developments in multi-modal LLMs \cite{table1-5} and specialized agents \cite{table1-6} provide added capability in financial sentiment analysis and market predictions, underscoring the significant role of LLMs in shaping the future of financial analytics. Table \ref{tab:ex} summarizes the recent research on LLMs in the financial field, detailing their contributions.
\begin{table*}[ht] 
\caption{Existing algorithms developed for Financial insights, trends and assessments.}
\scalebox{0.8}
    {
\begin{tabular}{|l|c|l|}
\hline
 \textbf{Paper Title}&\textbf{Year}& \textbf{Core Capabilities}\\ \hline
 FINBERT: A Large Language Model for&2022&- It is a LLM that adapts to the financial domain being trained using Google’s BERT algorithm.\\
Extracting Information from Financial Text\cite{table1-1}&&- It is trained on a large corpus of unlabeled financial texts including corporate filings,\\
&&analyst reports, and earning conference call scripts.\\\hline
What do LLMs know about Financial Markets?&2022&- The LLM is prompted to produce Chain-of-Thought summaries to produce labels for financial sentiment.\\
A Case Study on Reddit Market Sentiment Analysis\cite{table1-2}&&- It is tested with GPT 3/ PALM to understand market sentiment.\\\hline
Data-centric FinGPT: Democratizing Internet-scale&2023&- This open-source framework has collected and curated financial data from 30+ diverse online sources.\\
Data for Financial Large Language Models\cite{table1-3}&&- The use-cases includes advisory, sentiment analysis, low-code development.\\\hline
Bloomberg GPT: A Large Language Model for Finance\cite{bloomberg}&2023&-This 50-billion parameter LLM is trained on a wide range of financial data.\\
&&-It is trained on 363 billion token datasets constructed based on Bloomberg’s data sources\\ 
&&-Training data is augmented with 345 billion tokens from general purpose datasets.\\\hline
Modal-adaptive Knowledge-enhanced Graph-based Financial&2024&-Video features, audio features, and text features (multi-modal information) are used to predict\\
Prediction from Monetary Policy Conference Calls with LLM\cite{table1-5}&&price movement and volatility by understanding the monetary policy conference calls.\\
&&BEiT-3 and Hidden-unit Bert (HuBERT) used to extract video\\ 
&&and audio features and ChatGLM2 as for processing text features.\\\hline
Designing Heterogeneous LLM Agents for &2024&-This instantiates specialized agents using prior domain knowledge of errors \\
Financial Sentiment Analysis\cite{table1-6}&&made in financial sentiment analysis (FSA).\\
&&-Agent discussions helped improve accuracies for FSA without needing to fine-tune the LLM model.\\ \hline
\end{tabular}\label{tab:ex}
}\vspace{-0.3cm}
\end{table*}

Financial report data is typically characterized by a mix of structured (tables, charts) and unstructured (narrative text) information. Here, techniques like Named Entity Recognition (NER) are used to extract key financial metrics from these texts. These reports are often lengthy and complex, with hidden data labels that require careful analysis to assess model performance. Additionally, financial reports contain important sentiment and tone information, crucial for financial analysis and transparency. Table \ref{tab:ex1} provides a summary of recent studies highlighting these specific characteristics of the finance-domain reports and data.
\begin{table*}[ht] 
\centering
\caption{Characteristics of financial report data.}
\scalebox{0.99}
    {
\begin{tabular}{|l|c|l|}
\hline
 \textbf{Paper Title}&\textbf{Year}& \textbf{Domain-specific findings}\\ \hline
Comprehensive Review of Text-mining &2020&-Financial reports contain a mixture of the following:\\
applications in Finance\cite{table2-1}&& structured data (tables and charts) and unstructured narratives\\
&&-NER using standard libraries (nltk, spacy) can be used to extract \\
&&entities (e.g., company names, financial metrics) from unstructured texts.\\ 
&&-Context-dependent (domain specific) language can be often found\\ 
&&in financial reports like annual reports.\\ \hline
GPT-InvestAR: Enhancing Stock Investment Strategies&2023&-To assess the LLM performance, domain-specific data labeling is required\\
through Annual Report Analysis with LLMs\cite{table2-2}&& to report accuracy. E.g.: ``\textit{percentage return} of each stock between \textit{filing dates}''.\\ \hline
NLP Sentiment Analysis and Accounting&2023&-Financial reports contain important sentiment/tone and sentence formation \\
Transparency: A New Era of Financial Record Keeping\cite{table2-3}&& information that needs to be maintained across paragraphs.\\ \hline
\end{tabular}\label{tab:ex1}
}\vspace{-0.3cm}
\end{table*}

\subsection{LLMs for style-transfer}
Research for writing style-transfer has seen significant development in recent years as shown in Table \ref{tab:ex2}. Initial efforts focused on creating datasets and methodologies for formality style transfer\cite{table3-1}, with advancements leading to more sophisticated techniques for maintaining semantic integrity while altering style. By 2023, models like ChatGPT demonstrated improved capabilities in evaluating and editing text for style transfer\cite{table3-5}. Recent research explored multidimensional evaluations and integrated new approaches for enhancing style adaptation. However, there is a need for methods that enable effective style-transfer in the financial domain to address the content understanding (tables to answers) challenge and to generate semantically-acceptable text with minimal prompting/instructions to conserve input token size constrains for LLMs. This work is aimed to address this need to develop financial text generation methods with minimal data and fine-tuning costs.
\begin{table*}[ht] 
\caption{Examples of Style Transfer works.}
\scalebox{0.78}
    {
\begin{tabular}{|l|c|l|}
\hline
 \textbf{Paper Title}&\textbf{Year}& \textbf{Core Capabilities}\\ \hline
Dear Sir or Madam, May I Introduce the GYAFC Dataset:&2018&-Parallel supervision: source-target sentence pairs are labeled.\\
Corpus, Benchmarks and Metrics for Formality Style Transfer\cite{table3-1}&&-Early work on style transfer with large corpus of training data.\\ \hline
Disentangled Representation Learning for Non-Parallel&2019&-Non-parallel supervision with style labels.\\
Text Style Transfer\cite{table3-2}&&-Latent representations of style and content need to be learned separately.\\ \hline
ChatGPT vs Human-authored Text: Insights into Controllable&2023&-Text style transfer can be summarized as the task that involves transforming an input text\\
Text Summarization and Sentence Style Transfer\cite{table3-3}&&to a target style while maintaining the style-independent semantics.\\ 
&&-To assess ChatGPT’s summarization performance, the following metrics are used: Flesch Reading Ease,\\
&& Coleman-Liau Index (CLI), Dale-Chall Readability Score (DCR), Rouge Score; to assess formal style,\\
&&the following metrics are used: Formality Indicator, MTLD Lexical Diversity metric.\\
&&-Evaluation shows stylistic variations produced by humans are considerably larger than\\
&&those demonstrated by ChatGPT.\\ \hline
Prompt-based Editing for Text Style Transfer\cite{table3-4}&2023&-Prompt-based editing approach to text style transfer.\\
&&-Prompt a LLM for style classification and use classification probability to compute a style score.\\
&&Perform discrete search with word-level editing to maximize score function for style-transfer tasks.\\ \hline
Multidimensional Evaluation for Text Style Transfer\cite{table3-5}&2023&-Leveraged ChatGPT to evaluate the text transfer capabilities\\
using ChatGPT&& of other text style transfer models in the domain of: style strength, content preservation, and fluency\\
&&-ChatGPT achieves competitive correlations with human judgements\\ \hline
CAT-LLM: Prompting Large Language Models with Text Style Definitions\cite{table3-6}&2023&The model incorporates a text style definition module to comprehensively\\
for Chinese Article-style Transfer&&analyze text features in target articles from both words and sentences levels to learn target style.\\
&&-Evaluated with 5 styles of Chinese articles.\\ \hline
Whose LLM is it Anyway? Linguistic Comparison and LLM&2024&-Linguistic styles between three popular models are analyzed\\
Attribution for GPT-3.5, GPT 4 and Bard\cite{table3-7}&& in terms of: vocabulary, Part-Of-Speech distribution, dependency distribution, sentiment.\\
&&-The results point to significant linguistic variations.\\ \hline
Unsupervised Text Style Transfer via LLMs&2024&-Attention masking and LLM models are effectively combined to support unsupervised\\
and Attention Masking with Multi-way Interactions\cite{table3-8}&& text style-transfer in this paper.\\ \hline
\end{tabular}\label{tab:ex2}
}\vspace{-0.3cm}
\end{table*}

\subsection{Hallucination and Creativity in LLMs}
While fine-tuning LLMs for domain-specific writing styles such as finance, sales, medical reporting etc. are necessary, most LLM fine-tuning techniques do not focus on hallucinations introduced by un-prepared data sources. A recent work in \cite{hall1} presents the mathematical framework to define LLM hallucinations using probability and information theoretic approaches. This work demonstrates that LLM hallucinations are characterized by low probabilities of sequential tokens. Also, it illustrates that hallucinations arise from self-supervised learning since the training process typically relies on metrics such as ROUGE, TER, BLEU \cite{metrics} etc. that focus on ensuring that the response stay similar to the context, even if a well formed answer already exists. Also, hallucinations are \textit{implausible} based on the context and can be considered as inference-level anomalies that cannot be replicated owing to low token probabilities. In this work, we expand on these observations and utilize metrics such as averaged sequential loss per sentence (ASLS) and cross entropy (CE) loss to detect the likelihood of sequential token generation across training epochs. 

Another recent work in \cite{hal2} demonstrates the two phases of LLM hallucinations, wherein, the first divergent-phase hallucination induces creativity that can be controlled by advanced prompt engineering, and fine-tuning to promote creativity. The second convergent-phase hallucination involves standard RAG scenarios \cite{bigdata} that require intention recognition and hallucination detection through RAG-based control mechanisms. This work expands on the divergent phase to carefully pre-process the data followed by hallucination control flags to detect creativity.

\section{Methods and Data}
LLMs typically perform two basic tasks of natural language understanding (NLU), wherein the user data and query is converted to machine translation entities or tokens followed by natural language generation (NLG), wherein a sequence of words/tokens are generated based on the probabilities of the prior generated words. In this work, we focus on the NLG aspect of an LLM and the ability to transfer domain specific jargon, such as compound sentence generation and creative language generation. To evaluate the LLM responses, we analyze the user-query ($Q$), contextual data ($D$) and the LLM responses ($R$) together.

From prior works \cite{hall1} \cite{hal2}, we know that hallucinations and creativity in generated tokens have similar model-level level characteristics. Therefore, by ensuring a low value of top\_p and temperature, both hallucinations and creativity can be considerably reduced \cite{halu}. However, in this work, our goal is to minimize hallucinations while promoting creative sentence generation. As an example, consider the following data context, queries and responses $R_1, R_2$: 
\begin{itemize}
\item $D$: ``The company ACL had targeted 30\% profits but it finished Q2 at 28.8\% profits."
\item $Q$: ``How was ACL's performance in Q2?".
\item $R_1$: ``ACL met its target of 30\% profit in the Q2 quarter.''
\item $R_2$: ``ACL missed the planned target of 30\% by 1.2\% by the close of Q2."
\end{itemize}
In this situation $R_1$ is a \textit{hallucination} while $R_2$ is a \textit{creative} response. We assess the \textit{quality} of generated sentences after domain-specific fine tuning (FT) to isolate the log-probabilities at sentence level for creative versus hallucinated sentences. Additionally, the entities (nouns, locations, currencies etc.) in the generated text ($e_k$) and their relationships ($\rho_k$) can be extracted using standard libraries such as ``nltk'' and ``spacy'' to asses the formation of compound sentences in terms of the density of entities and relationships per sentence. The metrics used to evaluate the \textit{quality} of fine-tuned text are shown in section \ref{met}. For our experiments, we perform two-stage FT on the OpenAI GPT3.5 model using the RLHF technique \cite{openai}. As training data, we collect public domain financial reports that are labeled for financial entities and pre-processed into the ``prompt-completion" format as shown in section \ref{dataprep}. 

\subsection{Notation and Metrics}\label{met}
The goal of this work is to generate domain-friendly natural language text from a minimal prompt that contains basic instructions and financial data in a tabular format. For a sequence of words/tokens, the $i$'th generated token is $x_i$ and the log probability associated with the token is $p_i$. It is noteworthy that each sequential token is generated as a function ($F$) of the prior tokens and the token with highest probability across the top contenders for the $i$'th position ($x_i$), represented by equation \eqref{eq1}. Also, the log-probability of top $n$ contenders for each sequential position is collected as $P_i$ from the output of OpenAI's GPT3.5 to quantitatively detect creativity and hallucinations.
\begin{equation}\label{eq1}
    x_i=F(x_{i-1},x_{i-2}...,\arg\max(P_i)), \forall P_i=\{p_{i,1},p_{i,2}..p_{i,5}\}.
\end{equation}
As an example, each response word in the second column in Table \ref{tab:tgen} is selected across top $n=5$ contenders, as the word/token one with the highest probability (or lowest log-probability). The value of $n$ is selected empirically based on the relative distribution of log probabilities and the observation that for most generations, the log probability of tokens drastically reduces after 5.

Further, we assess the quality of fine-tuned generated text using the sequential log-probabilities per token and the following metrics.
\begin{itemize}
    \item Perplexity ($Per$): A lower value signifies that each sequential token is generated with high confidence following the FT process in equation \eqref{2}. $t$ represents the number of tokens per sentence.
    \item BLUE (Bilingual Evaluation Understudy) score \cite{metrics}: A high value represents high similarity between the generated text to the reference context ($D$), thereby representing accuracy of the generated text.
    \item TER (Translation Edit Rate) \cite{metrics}: A lower value indicates fewer edits required to transform the generated text to the reference context, thereby representing higher quality of generated text.
    \item ROUGE (Recall-Oriented Understudy for Gisting Evaluation) score \cite{metrics}: A high value represents high similarity between generated text and reference context.
    \item chrF++ (character level F score) \cite{metrics}: A high value operates at a character level to denote accuracy of generation in terms of similarity with reference context.
    \item Averaged sequential log-loss per sentence (ASLS): A high value represents highly discernible tokens in a sentence, given $t$ tokens per sentence. High ASLS is indicative of a highly non-uniform probability distribution per token $P_i$ and is shown in equation (3). This novel sentence-level metric evaluates how confident the LLM is in generating subsequent tokens. For instance, if top $n$ log-probabilities per sequential tokens are equally likely, that would result in a low ASLS, which indicates un-learned or hallucinated or creative text generation. ASLS is an extension from cross entropy loss ($CE$ in equation (4)), where a lower value indicates higher confidence of token selection. 
    \item Knowledge density per sentence (KDPS): A high value indicates dense information in terms of entities ($e_k$) and their relationships ($\rho_k$) per sentence, given $s$ sentences in a paragraph. This is a domain-specific metric and is representative of compound sentences at paragraph level in equation (5).
\end{itemize}
The impact of the well-known language evaluation metrics perplexity, BLUE, TER, ROUGE and chrF++ on LLM generated responses varying in certainty is illustrated by the examples in Table \ref{tab:tgen}. Here, we observe that for relatively close-ended user queries, metrics ($Per$, TER) have considerably low whereas (BLUE, ROUGE and chrF++) have higher values. Our goal with two-stage FT process is to ensure similar qualities in generated text from simple prompts and tabular contextual data for paragraph level financial report generation.

\begin{align}\label{2}
    Per=\exp(-\frac{1}{t}\sum_i^t (x_i|x_{i-1...})),\\
    ASLS=-\frac{1}{t}\sum_i^t\sum_j^n p_{i,j},\\
    CE=\sum_i^t-{\max(p_{i,j})_{j=1}^{n}},\\
    KDPS=\frac{1}{s}\sum_{k=1}^s (e_k+\rho_k).
\end{align}

\begin{table*}[ht] 
\caption{Examples of text generation Scores on well known user queries.}
\scalebox{0.65}
    {
\begin{tabular}{|c|l|c|}
\hline
 \textbf{User-Query}&\textbf{Response}& \textbf{Metrics}\\ \hline
 Was Constantinople officially renamed Istanbul in 1930?&Yes, Constantinople was officially renamed Istanbul in 1930 by the Turkish government.&$Per$=1.03, BLEU=27.82, ROUGE=0.63, chrF++=78.16, TER=100\\ \hline
 Did the Ottoman Empire capture Constantinople in 1453?&Yes, the Ottoman Empire captured Constantinople in 1453, marking the end of the  &$Per$=1.16, BLEU=4.3, ROUGE=0.24, chrF++=44.93, TER=550.0\\ 
 &Byzantine Empire and the beginning of Ottoman rule in the region.&\\\hline
How does Istanbul's unique geographical position, between &Istanbul's unique geographical position as a city that straddles Europe and Asia&\\
Europe and Asia, affect its role in international trade and diplomacy?&has historically made it a key hub for international trade and diplomacy. ....&\\
&Overall, Istanbul's unique geographical position as a city continues to shape its role &$Per$=1.2, BLEU=5.01, ROUGE=0.16, chrF++=33.48, TER=952.63\\
&in international trade and diplomacy today.&\\ \hline
\end{tabular}\label{tab:tgen}
}\vspace{-0.3cm}
\end{table*}

\subsection{Data Preparation and two-stage FT-process}\label{dataprep}
For our two-stage FT process, we generate prompt-response pairs from public-domain financial reports. The first step here is to generate sample data tables that can serve as inputs. For this, we generate a simple prompt that takes 150 financial report text as input and outputs the data in tabular format. This tabular data is further augmented by small variations to the numbers and schema through minimal LLM prompt alterations and manually verified for accuracy. This resulted in 1000 samples of tabular data and their adjoining reports. Next, we begin the two-stage FT process.
 
\begin{figure}[ht]
     \centering
    \subfigure[Data preparation in first stage.]
	{\includegraphics[width=0.45\textwidth]{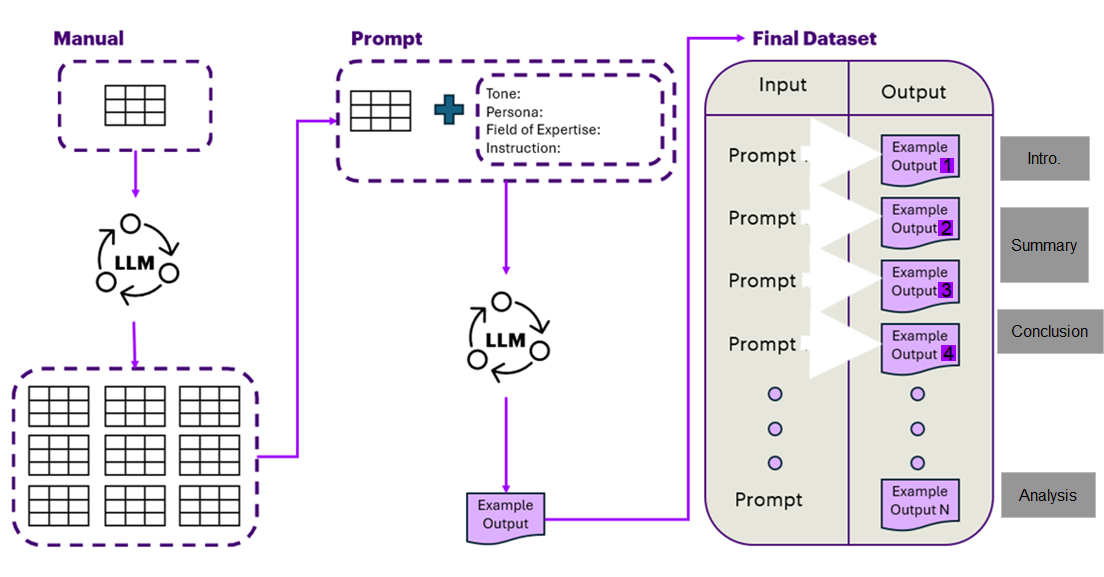}}
	\subfigure[Prompts for second stage error-correction]
	{\includegraphics[width=0.55\textwidth]{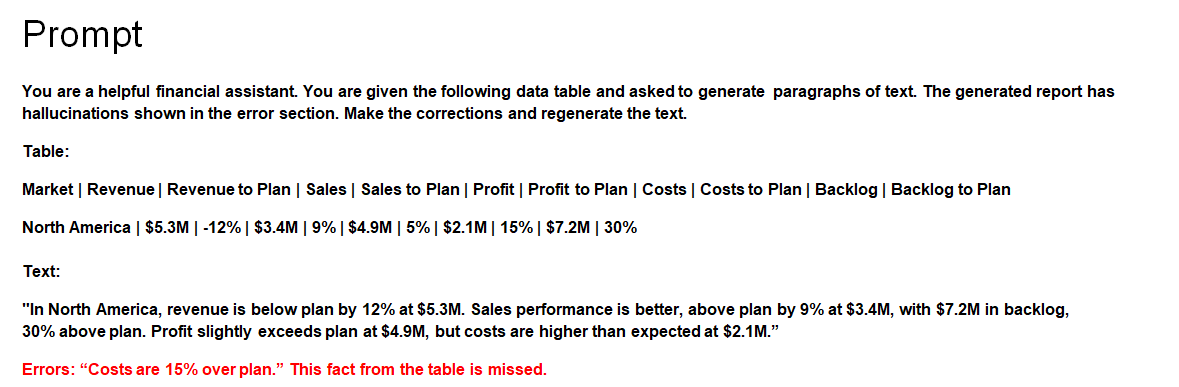}}
 	\subfigure[Completions for second stage error corrections.]
	{\includegraphics[width=0.55\textwidth]{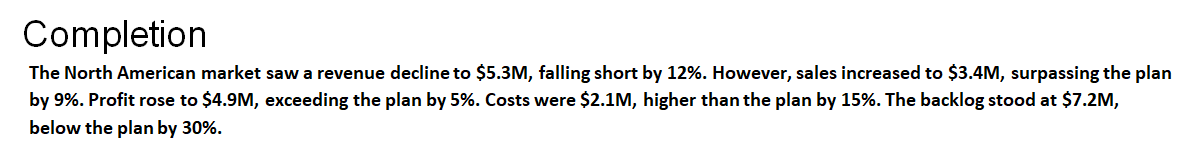}}
    \caption{Data Preparation process for the two-stage FT process.}
    \label{fig:dprep}
\end{figure}

For the first FT step, we generate a variety of sections from the financial reports such as \{introduction, analysis, conclusion, discussion\} sections. Each section requires specific verbiage, prompt-completions, specific instructions and examples. The prompt aspects that remain unchanged across all the prompt-completion samples are style-transfer attributes like tone, assertiveness, and persona. Thus, with minimal changes to a GPT4o LLM prompt, we obtain several prompts and their completions (expected outputs) that correspond to a variety of financial report sections as shown in Fig. \ref{fig:dprep}(a). 

The second error-correction step of the FT process involves manual detection of hallucinations, incomplete sentences and poor quality sentences resulting from feeding the above pre-processed data to GPT3.5 for 100 epochs while monitoring for [$Per, TER, BLUE$] metrics. Samples of \textit{poor quality sentences} and hallucinations are manually corrected, and from these corrections, we generate 800 samples of error-correction prompts as shown in Fig. \ref{fig:dprep}(b) and their corresponding completions as shown in Fig. \ref{fig:dprep}(c). Additionally, for end-to-end validation, two 10-page detailed financial reports are manually annotated for verbiage and quality of sentences.

\section{Experiments and Results}
We perform three sets of experiments to assess the \textit{quality} of generated text after the proposed two-stage FT process. First, we analyze the average $Per$ per generated section in comparison to the manually annotated reports using out-of-the-box (vanilla GPT3.5) model, single stage FT model (that is fine-tuned for style), and two-stage FT model (that is fine-tuned for hallucination correction). Second, we qualitatively evaluate the generated financial sections to explain the metrics for each model at a sentence level per generated paragraph. Third, we track the NLG metrics from section \ref{met} to analyze the quality of generated financial report sections across all the generated sample paragraphs and sections.
\subsection{Validation and Test Prompts}
The $Per$ for the two-stage FT model on the two manually validated reports is shown in Table \ref{tab:r1}. Here, we observe that the FT $Per$ are consistently lower and hence better per section of the financial reports.
\begin{table}[ht] 
\caption{Average Sectional perplexity in fine-tuned validation reports.}
\scalebox{0.9}
    {
\begin{tabular}{|c|c|c|}
\hline
\textbf{Report 1}&&\\
\textbf{Section}&\textbf{vanlillaGPT3.5 $Per$}&\textbf{Two-step FT, $Per$}\\ \hline
Introduction&1.58&1.25\\
Growth Outlook&1.24&1.21\\
Service Group Performance&1.26&1.26\\
Industry Performance&1.35&1.33\\
Performance Highlights& 1.57&1.51\\ \hline
\textbf{Report 2}&&\\ \hline
Introduction&1.245&1.21\\
Financial Review& 1.186&1.08\\
New Bookings& 1.225&1.07\\
Revenues by Geographic Market& 1.083&1.07\\
Revenues by Industry Group& 1.034&1.005\\
Returning Cash to Shareholders& 1.205&1.15\\
Business Outlook& 1.163&1.109\\ \hline
\end{tabular}\label{tab:r1}
}\vspace{-0.3cm}
\end{table}
Additionally, the average scaled KDPS for vanilla GPT3.5 and two-step FT model on both reports are 0.75 and 0.8 respectively. This demonstrates compound sentences and increased number of entity relations per sentence in the FT model.

\begin{figure}[ht]
    \centering
    \includegraphics[width=3.4in]{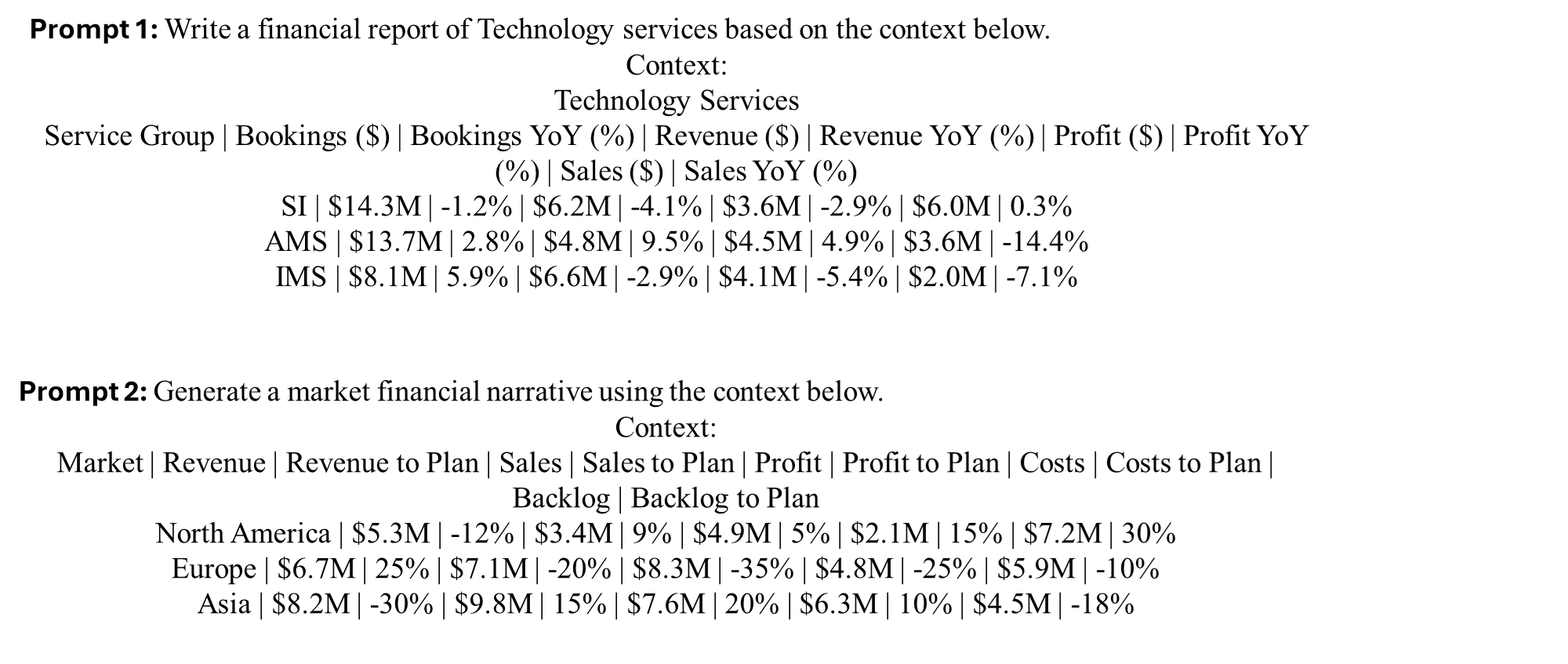}
    \caption{Prompt 1 and 2 used for testing performance of financial reports}
    \label{fig:p12}
\end{figure}

\begin{figure}[ht]
    \centering
    \includegraphics[width=3.4in]{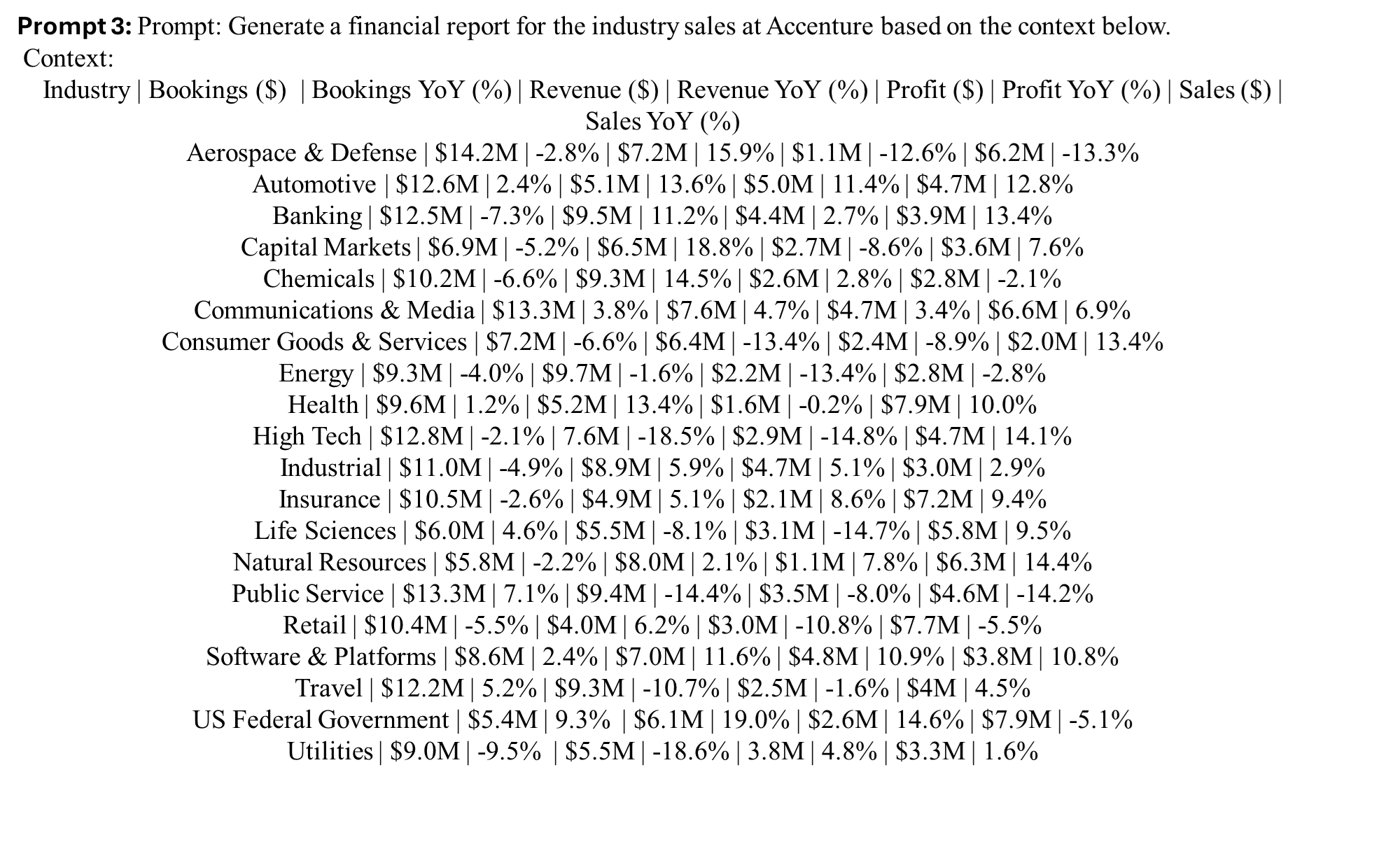}
    \caption{Prompt 3 used for testing performance of financial reports}
    \label{fig:p3}
\end{figure}

Next, we assess the report generation performances for the following 3 prompts shown in Fig. \ref{fig:p12} and Fig. \ref{fig:p3}.

\subsection{Quantitative Analysis of Two-stage FT model}
\begin{figure}[ht]
    \centering
    \includegraphics[width=3.3in]{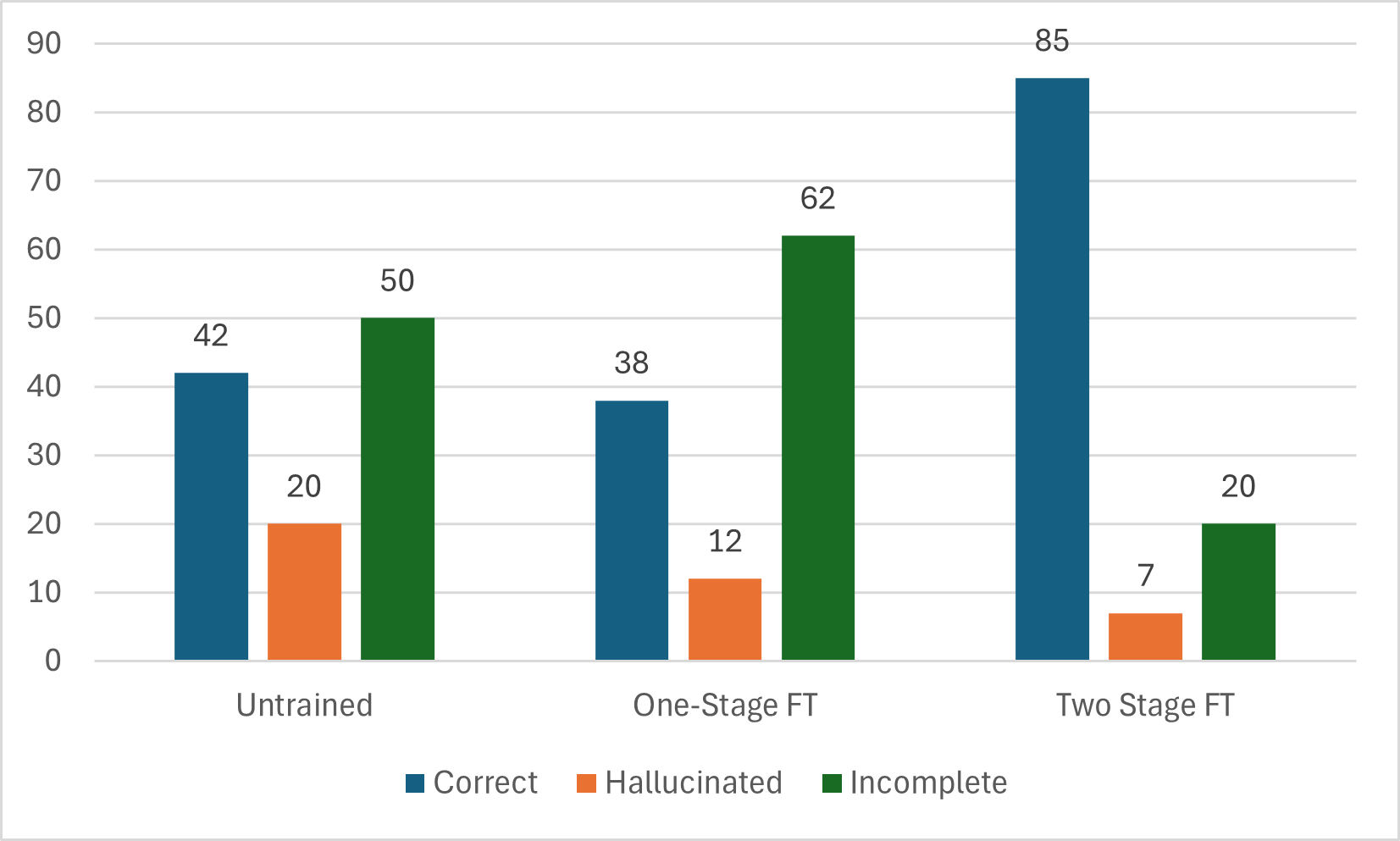}
    \caption{Question answering performances on the untrained, one-stage and two-stage FT models.}
    \label{fig:R2}
\end{figure}

To ensure style transfer and hallucination control, the one-stage FT model (style only) involves fine-tuning vanilla GPT3.5 model with 1000 prompt-completions generated using the method explained in section \ref{dataprep} over 100 epochs. Next, we query 112 questions to the one-stage FT model and check the performance for hallucinations. For the two-stage FT model (style and hallucination) an additional FT process involves 800 samples of hallucinations that are manually corrected and subjected to further tuning for 100 epochs. The quality of responses to the 112 user queries can be categorized as \{Correct, Hallucinations, Incomplete\} and shown in Fig. \ref {fig:R2}. We observe that the two-stage FT model doubles correct answering capability (increase from 42\% to 85\%) and significantly reduces hallucinations (reduction from 20\% to 7\%) when compared to an untrained GPT3.5 model. 

\begin{figure}[ht]
    \centering
    \includegraphics[width=3.3in]{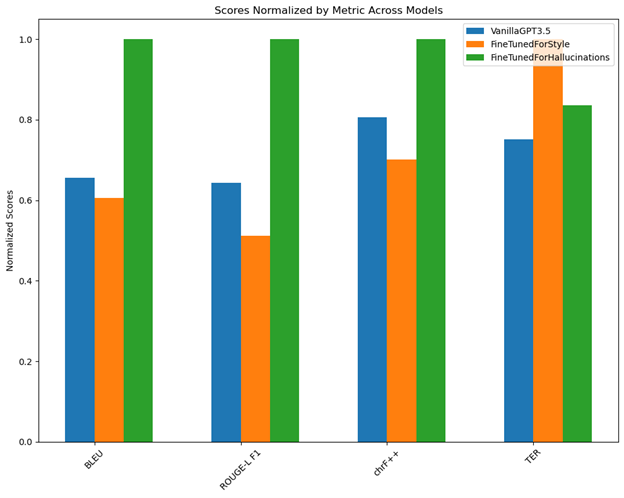}
    \caption{Quality of generated text for the untrained, one-stage and two-stage FT models.}
    \label{fig:R1}
\end{figure}

Additionally, the quality of generated text is further shown in terms of scaled metrics in Fig. \ref{fig:R1}. Here we observe that the two-stage FT model has consistently high BLEU, ROUGE and chrF++ metrics when compared to the untrained and one-stage FT models.

\subsection{Qualitative Analysis: Hallucination, Creativity monitoring}
\begin{figure*}[ht]
    \centering
     \subfigure[Generated text by vanilla GPT3.5 model. Low information sentences are in red.]
	{\includegraphics[width=0.9\textwidth]{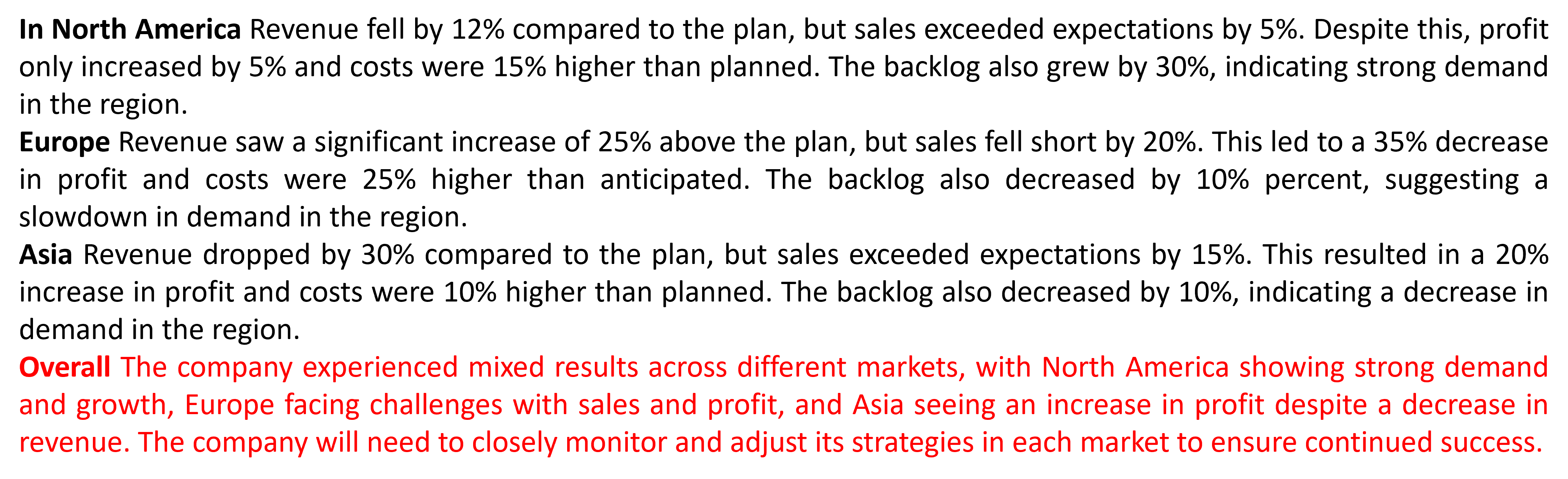}}
	\subfigure[Generated text after one-stage FT process. Low information sentences are in red.]
	{\includegraphics[width=0.9\textwidth]{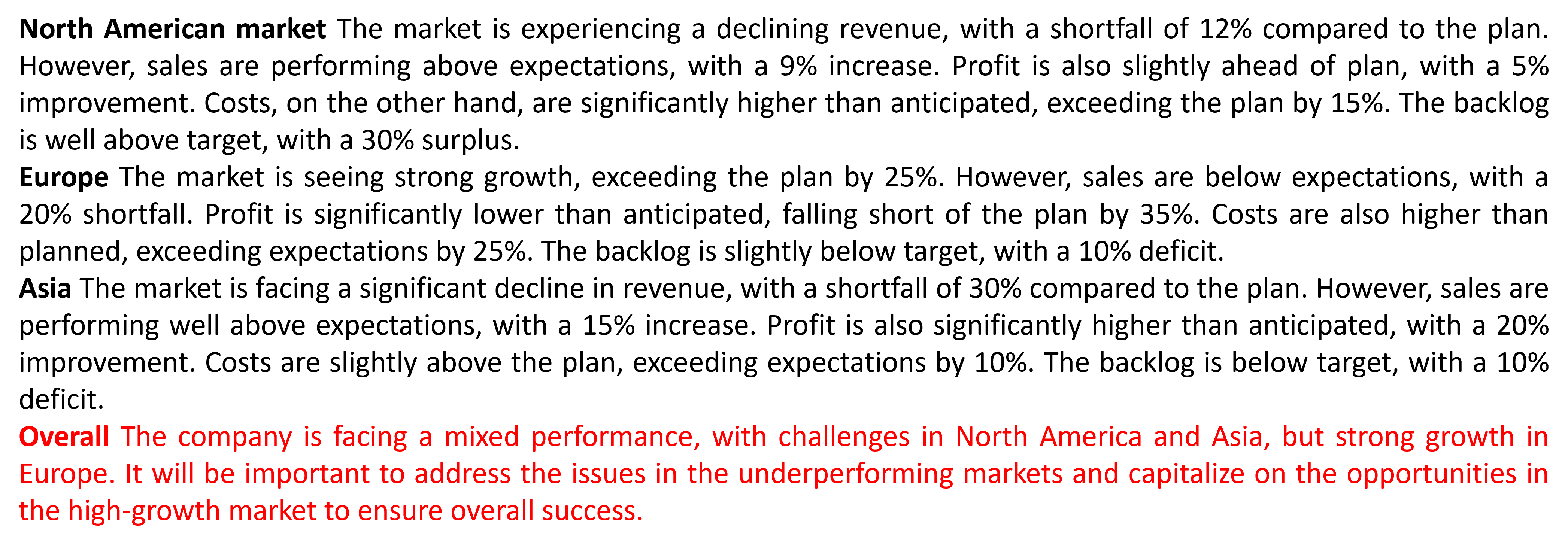}}
 	\subfigure[Generated text after two-stage FT process.]
	{\includegraphics[width=0.9\textwidth]{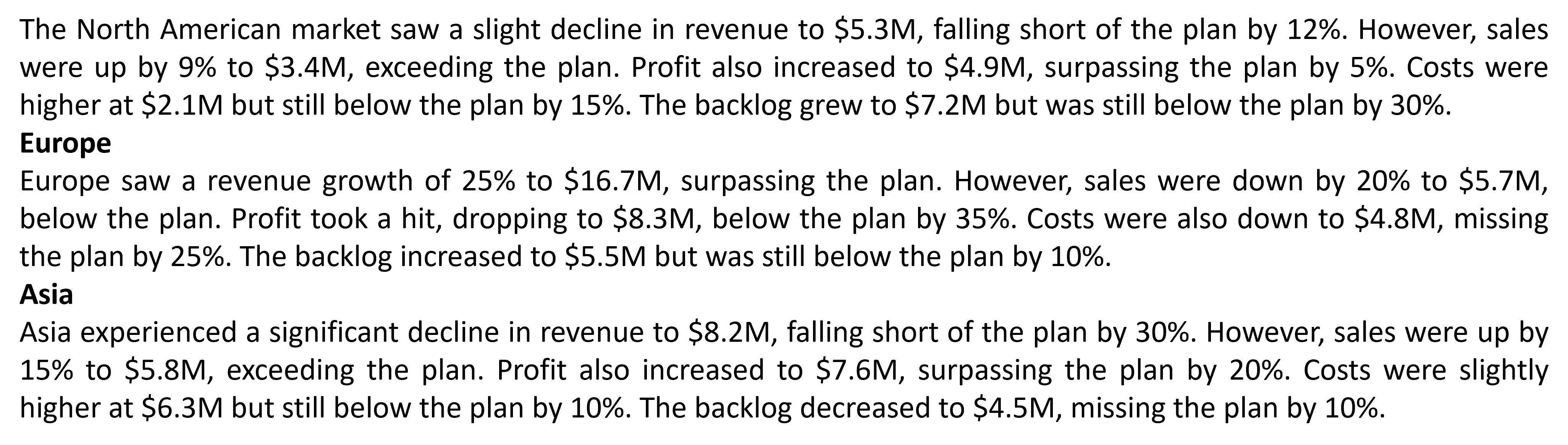}}
    \caption{Examples of untrained, one-stage FT and two-stage FT outcomes. The last two lines from the vanilla GPT3.5 and one-stage FT model have minimal information and entities. Our goal is to reduce such sentences with low information content.}
    \label{fig:explain}
\end{figure*}

In the previous subsection, we observe the importance in the two-stage FT process to control for hallucinations and incomplete sentences. However, since creative and hallucinated words/tokens have similar probabilistic nature, we qualitatively assess the reports generated from the prompts shown in Fig. \ref{fig:p12} and Fig. \ref{fig:p3} for hallucinations and creativity. Fig. \ref{fig:explain}(a), (b), (c) demonstrates sample reports generated from vanilla GPT3.5, one-stage FT and two-stage FT models, respectively.

\begin{figure*}[ht]
    \centering
    \subfigure[Scatter plot for cross entropy loss per sentence after each stage of fine-tuning.]
    {\includegraphics[width=0.45\textwidth]{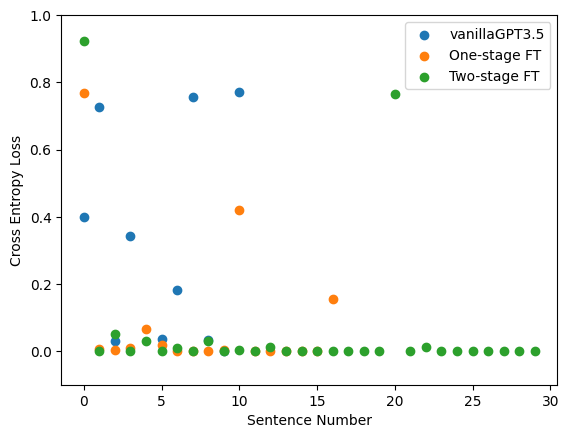}}
    \subfigure[Scatter plot for ASLS before and after each stage of FT.]
    {\includegraphics[width=0.45\textwidth]{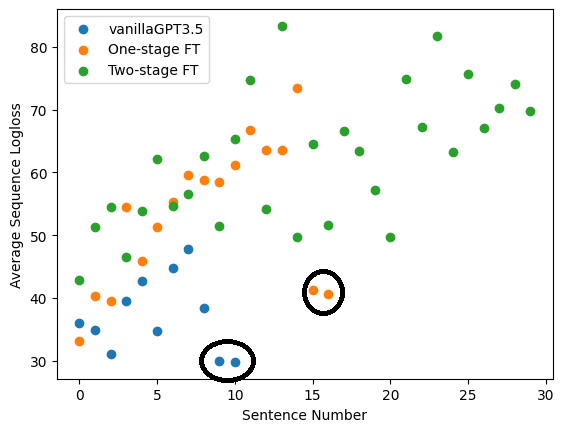}}
    \caption{}
    \label{fig:ce}
\end{figure*}

 In Fig. \ref{fig:ce} we observe that the cross-entropy of the generated text significantly reduces after each stage of FT leading to less hallucinations and more certain text generation. 

Fig. \ref{fig:ce} shows the ASLS metrics at sentence level for the vanilla GPT3.5, one-stage FT and two-stage FT models, respectively. We observe that for the untrained and one-stage FT models, the last two lines have the lowest log-loss, or low certainty for text generation. These sentences have a higher tendency to contain hallucinations or creativity. However, for the two-stage FT model, the hallucinations significantly reduce, thereby leading to more certain and higher quality of text generation.

\section{Conclusions and Discussion}\label {conclusion}
In this work we present a two-stage LLM fine tuning process, starting from data pre-processing to monitoring metrics to enable financial report generation that is similar in style to a financial analyst. Our goal is to minimize domain-specific fine tuning costs and explore data processing methods to enhance self learning and minimize hallucinations from LLMs. Our generalizable two-stage FT process incurs \$18 costs for the final version of fine-tuned GPT3.5 LLM. The data pre-processing steps are augmented with minimal GPT4o prompts and the two-stage FT process doubles the correct response rate and halves the rate of hallucinations and incomplete responses. This makes the proposed two-stage FT model the preferred choice for generating accurate, coherent, and appropriately detailed financial report generation use-case.

The novelty of this work lies in the nature of the FT process, since we allow hallucinations in the first stage. In the second stage the hallucinations are corrected and the LLM is allowed to self-learn from the corrections. This process enhances the creativity and compound sentence generation capabilities of the LLM, enabling domain-specific fine-tuning at low costs. It is noteworthy that analysis on a variety of versions of LLM models on financial reports reveal the following nature of generated text.
\begin{itemize}
    \item Closed-ended questions generally have lower $Per$, higher ROUGE and lower TER scores indicating that they are easier for the model to predict based on reference context.
    \item Open-ended questions tend to have lower BLEU, higher TER and perplexity suggesting more uncertainty in the generated text. 
\end{itemize}
This work aims to generate close-ended text quality for open-ended questions where the prompts include minimal instructions and tabular data as input for multiple paragraph generation tasks. Additionally, we observe that for the labeled validation reports, the fine-tuned models generally outperform the untrained model with higher average coherence and correctness, while the relevance of data remains fairly unchanged. For question-answering from minimal prompts, the FT models excel in depth and creativity, although creativity is not a desired trait for financial reports. However, untrained models typically demonstrate generic conciseness. Thus, the fine-tuned models demonstrate superior performance in coherence and correctness, combined with appropriate depth. This makes the proposed two-stage FT process the preferred choice for sectional financial report generation. Future works will be directed towards further controlling for hallucinations and creativity by monitoring weights and biases of the early layers of the LLMs.

\bibliographystyle{IEEEtran}
\bibliography{main}

\end{document}